\documentclass[11pt]{article}

\usepackage[preprint]{acl}

\usepackage{times}
\usepackage{latexsym}

\usepackage[T1]{fontenc}

\usepackage[utf8]{inputenc}

\usepackage{microtype}

\usepackage{inconsolata}

\usepackage{graphicx}

\usepackage{amsmath}

\usepackage{booktabs}

\title{Inspicio: Open-Vocabulary, LLM-Based Sense Retrieval for Historical Languages}

\author{Michele Ciletti \\
  Department of Humanities, University of Foggia / Via Arpi 176, 71121 Foggia, Italy \\
  \texttt{michele.ciletti@unifg.it} \\
}

\begin{document}
\maketitle
\begin{abstract}
Word Sense Disambiguation has advanced rapidly for English and a handful of well-resourced modern languages, but it continues to assume the existence of a sense inventory and a word-to-sense mapping in the source language \citep{navigli2026word}. These assumptions break down for most historical and low-resource languages, whose dedicated WordNets are either incomplete or still under construction. We present \textsc{Inspicio}, an open-vocabulary retrieval pipeline that links tokens in context to synsets of the Open English WordNet \citep{mccrae2020english} without requiring any source-language inventory or mapping. For each occurrence, an instruction-tuned LLM produces two English translations of the surrounding sentence, a small set of candidate dictionary-style definitions, and a few candidate English lemmas. These outputs drive a hybrid retrieval step that combines dense definition–synset similarity, sparse lemma matching, and Maximal Marginal Relevance re-ranking. We evaluate the pipeline across a $6 \times 6$ grid of LLMs and sentence-embedding models on a new bilingual set of manually annotated Latin and Ancient Greek perception verbs, on the PREMOVE dataset\citep{farina2025premove}, and on a diachronic sample of Italian. The best configuration reaches 96\% Recall@50 on the perception-verb test set, with each component contributing measurable gains, and remains competitive in the out-of-domain and cross-lingual settings.
\end{abstract}

\section{Introduction}

Word Sense Disambiguation (WSD) has made considerable progress over the past decade, in particular
thanks to contextualized representations and Large Language Models
\citep{navigli2009word, bevilacqua2021recent, bejgu2024word, navigli2026word}.
Standard formulations of these tasks share a set of strong infrastructural
assumptions: (i) a sense inventory is available for the language under
analysis; (ii) the ambiguous spans to be disambiguated are either marked or
can be reliably identified; and (iii) a word-to-sense mapping lists the
candidate senses for each span. Under these conditions, modern systems
are able to reach high accuracy on English and a considerable number of well-resourced languages.

For most of the world's languages, and for nearly all historical ones, these
assumptions hold only partially. Comprehensive WordNets remain restricted to
a handful of contemporary languages, and the largest cross-lingual
evaluation framework for WSD currently covers only eighteen of them, all
modern \citep{pasini2021xl}. Building Wordnets by hand is
expensive; automatic transfer approaches, typically following the
\emph{expand} method \citep{vossen2004eurowordnet, pianta2002multiwordnet},
often inherit coverage gaps and noise from the source.

In this paper we present \textsc{Inspicio}, a retrieval pipeline that
performs open-vocabulary sense retrieval for historical and
low-resource languages without relying on a sense inventory in the source
language. Given a token in context, \textsc{Inspicio} prompts an LLM to
translate the surrounding sentence into English, to propose a small set of
candidate English definitions for the target word, and to suggest a few
English lemmas that could render it in that context. These outputs are then
used to query an embedding index built over the Open English WordNet
(OEWN, \citealp{mccrae2020english}). Retrieval is hybrid: dense similarity between LLM-generated
definitions and synset documents is combined with sparse, lemma-based
lookups, and the merged pool is diversified through Maximal Marginal
Relevance to counteract the well-known fine granularity of WordNet senses.
The choice of English as a pivot derives from the fact that the OEWN is the most
extensively populated lexical-semantic network of its kind
\citep{mccrae2020english}, and it is the same pivot
adopted by several bootstrapping projects that aim to create new Wordnets, including both the Latin and Ancient Greek WordNets
\citep{minozzi2017latin, bizzoni2014making}. The pipeline is otherwise
language-agnostic and runs in a fully zero-shot regime, requiring no
sense-annotated data in the source language and therefore being equally
applicable to other historical or under-resourced languages.

We evaluate \textsc{Inspicio} on a curated set of 150 perception-verb
tokens in Latin and Ancient Greek, manually annotated with gold OEWN
synsets, and additionally on PREMOVE \cite{farina2025premove}, a diachronic dataset of
preverbed motion verbs spanning eight centuries of Latin and Greek
literature. Finally, we test it against a small diachronic dataset of old and modern Italian to test its performance in an alternative setting. Our experiments are organized around one main and three
secondary research questions:

\begin{description}
  \item[\textbf{RQ1.}] Can an LLM-driven, gloss-guided retrieval pipeline
    produce a high-recall candidate pool of OEWN synsets for historical language tokens \emph{without} any source-language sense
    inventory or word-to-sense mapping?
  \item[\textbf{RQ2.}] How do the choice of generative LLM and the choice
    of sentence-embedding model interact in determining retrieval quality?
  \item[\textbf{RQ3.}] What is the relative contribution of dense
    (definition-based) and sparse (lemma-based) retrieval, and does their
    combination recover senses that either component would miss in
    isolation?
  \item[\textbf{RQ4.}] Does a pipeline tuned on a semantically focused case
    study generalize to a more heterogeneous, naturalistically distributed
    dataset and to new languages?
\end{description}


The code, prompts, and evaluation data described in this paper are
released on GitHub.\footnote{\url{https://anonymous.4open.science/r/inspicio}}

\section{Background and Related Work}

\subsection{LLM-based Word-Sense Disambiguation}

With the rise of large instruction-tuned LLMs, many researchers have asked whether these models, pretrained on broad text corpora, retain a fine-grained
understanding of word senses. Several studies have investigated this question
in zero- and few-shot regimes \citep{kocon2023chatgpt, yae2025leveraging,
kibria2024functional, capone2024lost}. A few consistent patterns are clear:
model scale correlates strongly with disambiguation accuracy; prompt
design has a non-negligible effect; and even the strongest systems are imperfect on rare or contextually subtle senses. \citet{meconi2025large}
evaluated a range of open- and closed-weight LLMs and found that top
systems such as GPT-4o approach the performance of specialized supervised
models, while still trailing expert human annotators. \citet{basile2025exploring}
reported analogous findings on a multilingual definition-selection
benchmark derived from XL-WSD, also showing that a fine-tuned medium-sized
LLM can surpass much larger zero-shot ones. Targeted prompting strategies
that inject glosses and elicit chain-of-thought reasoning yield further
gains \citep{sumanathilaka2025can}.

\subsection{Word-Sense Disambiguation for Historical Languages}

WSD for historical languages has long been constrained by sparse annotated
data, dictionary inventories of fine granularity, and limited cross-lingual
transferability. The first systematic study for Latin is the experiment of
\citet{bamman2020latin}, who fine-tuned Latin BERT on a binary sense
classification task derived from the Lewis and Short dictionary, restricted
to the two most frequent senses of each headword. \citet{lendvai2022finetuning}
extended this line of work with a larger sample of senses drawn from the
\textit{Thesaurus Linguae Latinae}.

The manually annotated Latin portion of the SemEval-2020 Lexical Semantic
Change dataset \citep{schlechtweg2020semeval, mcgillivray2022new} opened
the way to more elaborate strategies. \citet{ghinassi2024language}
propagated WSD annotations from English to Latin through a parallel
corpus, showing that automatically obtained labels can usefully complement
gold data, especially on under-represented senses. \citet{ghizzota2025meaning}
compared zero-shot and fine-tuned generative LLMs on the same resource
and confirmed that medium-sized fine-tuned models can match or exceed
much larger zero-shot ones. In a complementary direction,
\citet{ghizzota2026linguistic} integrated the dataset into a Linguistic
Knowledge Graph and tested Graph Retrieval-Augmented Generation for sense
prediction, finding that the benefit of structured authorial and lexical
metadata depends substantially on model size.

Dedicated studies for Ancient Greek are still few. Recent work has
explored corpus-based and transformer-based methods \citep{mercelis2025tongue}
and proposed semi-automatic pipelines for populating the Ancient Greek
WordNet through LLM-assisted annotation \citep{marchesi2025towards}.
Cross-linguistic experiments on Latin and Greek have addressed specific
lexical classes, including preverbed motion verbs \citep{farina2025probing}
and geographical common nouns \citep{farina2026sense}, with the latter
explicitly anchoring Latin and Greek tokens to English WordNet synsets
through LLM-driven annotation.

\subsection{Open-Vocabulary Word-Sense Disambiguation}

A separate strand of research has questioned the infrastructural
assumptions of classical WSD. \citet{bejgu2024word} formalised Word Sense
Linking as a task in which the spans to be disambiguated are identified
directly in raw text and linked to a reference inventory, without relying
on a pre-existing word-to-sense mapping. Their retriever-reader architecture
shows that it is feasible for English, but its performance degrades sharply when the source-language
mapping is incomplete—a situation that is the norm rather than the exception
for historical languages.

A few generative approaches have pushed this idea further by abandoning closed
inventories altogether. \citet{bevilacqua2020generationary} framed WSD as a gloss generation task. \citet{meconi2025large} subsequently
reported that LLMs explain word meanings substantially more accurately
when allowed to produce free-form definitions than when forced to select
from a fixed candidate list, reaching up to 98\% accuracy in their most
permissive setting.

\textsc{Inspicio}'s pipeline fits within this landscape by assuming neither a
source-language inventory nor a word-to-sense mapping; it relies on an LLM
to produce English-language hypotheses about meaning, and on a sentence
embedding model to align them with the Open English WordNet. The final
output is therefore expressed in terms of a stable and interpretable
lexical-semantic resource, while leaving the source side entirely
inventory-free.

\section{Methodology}
\label{sec:method}

\begin{figure*}[t]
    \centering
    \includegraphics[width=0.9\textwidth]{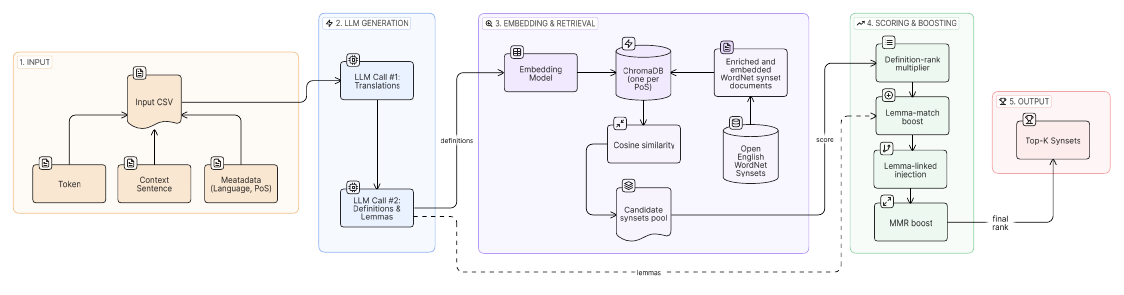}
    \caption{Architecture of the \textsc{Inspicio} pipeline, showing the transition from zero-shot translation to dense and sparse OEWN synset retrieval.}
    \label{fig:pipeline}
\end{figure*}

\textsc{Inspicio} processes one token at a time. Given a row consisting of
a target token, its dictionary lemma, the sentence in which it occurs, and
a language tag, the pipeline returns a ranked list of OEWN synsets,
together with the intermediate artifacts that produced the ranking.
Figure~\ref{fig:pipeline} summarises the architecture, which we describe
in the following subsections. Hyperparameter values are reported as we
introduce them; they were selected through iterative inspection of outputs
(Section~\ref{sec:data}).

\subsection{Translation and Hypothesis Generation}

The first two stages query a generative LLM in a fully zero-shot setting.
The first call asks the model to produce two English translations of the
input sentence: a \emph{literal} one, which stays close to source-language
word order and lexical choices, and a \emph{natural} one, which is fluent
in modern English. We use a relatively high temperature ($T = 1.0$, except for when the documentation of specific models explicitly requires a different value) to
encourage variation between the two outputs. This step creates a bridge to English
without committing to a single translation strategy: literal renderings
can preserve etymological cues that help recover compositional meanings,
whereas natural ones can often surface idiomatic readings that could
otherwise be lost.

The second call provides the LLM with the original sentence, the target
token, its lemma, and the two translations, and asks for two outputs:
between one and three English dictionary-style definitions of the target
word in context, ordered by likelihood, and between one and five candidate
English lemmas or short multiword expressions (e.g.\ \textit{make up},
\textit{look at}) that could render the token in that occurrence. The
prompt explicitly addresses well-known pitfalls of LLM-based sense
generation in light of recent findings \citep{meconi2025large}: it
instructs the model to define the verb itself rather than the
verb-plus-negation, to include both literal and metaphorical readings
when ambiguity is real, and to keep verb semantics distinct from argument
semantics. We lower the temperature ($T = 0.8$) so that the lexicographic
output remains stable while still allowing the model to surface multiple
plausible senses. Definitions and lemmas are returned as JSON.

\subsection{Embedding and Dense Retrieval}

We build four separate OEWN indexes \citep{mccrae2020english}, one for each part-of-speech covered by the OEWN itself (nouns, verbs, adjectives, adverbs), under the assumption that the target token has already been PoS-tagged. Our
experiments take into consideration the verb partition. Each synset is represented as a short
document concatenating its lemmas, gloss, examples, hypernyms, and
lexname; this representation outperformed gloss-only and lemma-only
variants in preliminary tests. The documents are encoded with a
sentence-embedding model (Section~\ref{sec:experiments} lists the models
compared) and stored in a ChromaDB collection. All embeddings are
$L_2$-normalised, so dot products coincide with cosine similarities.

For each LLM-generated definition $d$, we embed it with the same encoder
used for the synsets and retrieve its top $N = 50$ nearest synsets from
the index. The value of $N$ was chosen large enough to reliably include
the gold synset on a development subsample, while remaining adjustable for
downstream merging.

\subsection{Definition-Rank Decay Scoring}

A given synset may appear in the top-$N$ of more than one definition. We
aggregate its dense contributions through a weighted sum that respects
the order returned by the LLM:
\begin{equation}
\mathrm{S}_{\mathrm{base}}(s) \;=\; \sum_{d=1}^{D} w_d \cdot
\cos\!\left(\mathbf{e}_{d}, \mathbf{e}_{s}\right),
\end{equation}
where $D$ is the number of definitions returned (between one and three),
$\mathbf{e}_d$ and $\mathbf{e}_s$ are the embeddings of definition $d$
and synset $s$, and $w_d$ is a rank-decay weight. We set
$\mathbf{w} = (1.0,\, 0.75,\, 0.5)$, reflecting the prompt instruction
that the first definition encodes the model's most likely reading. The
similarity is taken only over those definitions for which $s$ entered
the top-$N$ pool; reusing the retrieval list, instead of re-encoding the
full pool against every definition, gave comparable rankings at a
fraction of the cost.

\subsection{Sparse Lemma-Based Retrieval}

In parallel with dense retrieval, the candidate English lemmas produced
in the second LLM call are used to query a precomputed inverted index that maps each
English lemma to the set of synsets containing it. This index is built
once from the ChromaDB metadata and cached on disk. The lemma pool
augments the dense one with synsets that may be semantically distant in
embedding space, but lexically anchored through translation. Such
synsets are common when the source-language token has an idiomatic
English counterpart that no single definition captures fully.

\subsection{Merging and Final Scoring}

The dense and sparse pools are merged into a single candidate set. Each
synset $s$ receives a final score
\begin{equation}
\mathrm{S}_{\mathrm{final}}(s) \;=\;
\begin{cases}
\mathrm{S}_{\mathrm{base}}(s) \cdot \left(1 + \gamma \cdot \mathbf{1}[s \in \mathcal{L}]\right) \\
\quad \text{if } \mathrm{S}_{\mathrm{base}}(s) > 0, \\[6pt]
\beta \\
\quad \text{if } \mathrm{S}_{\mathrm{base}}(s) = 0 \text{ and } s \in \mathcal{L},
\end{cases}
\end{equation}
where $\mathcal{L}$ denotes the lemma-matched set, $\gamma = 0.8$ is a
boost applied to synsets that are both densely and lexically supported,
and $\beta = 0.65$ is a fixed fallback score for synsets retrieved only
through the lemma index. The fallback prevents lemma-only candidates
from being silently discarded, while keeping them below the
dense-and-lemma combination in the ranking. The values of $\gamma$ and
$\beta$ were tuned on the development sample. The merged pool is
truncated to the top $500$ synsets by $\mathrm{S}_{\mathrm{final}}$
before the next stage.

\subsection{Diversification with Maximal Marginal Relevance}

OEWN exhibits well-known granularity issues, with closely related senses
often distinguished by subtle nuances \citep{bejgu2024word,
navigli2026word}. A relevance-only ranking therefore tends to fill the
top-$K$ with near-duplicate senses of the dominant reading. We re-rank
the merged pool with Maximal Marginal Relevance
\citep{carbonell1998use}:
\begin{equation}
\begin{split}
\mathrm{MMR}(s) \;=\; & \lambda \cdot \mathrm{S}_{\mathrm{final}}(s) \\
& -\; (1 - \lambda) \cdot \max_{s' \in \mathcal{S}} \cos\!\left(\mathbf{e}_s, \mathbf{e}_{s'}\right),
\end{split}
\end{equation}
where $\mathcal{S}$ is the set of already-selected synsets. We set
$\lambda = 0.8$, a relevance-leaning value that introduces enough
diversity to break clusters of near-identical senses while preserving
the top of the ranking. The final list is truncated to $K = 50$.

\subsection{Output and Auditability}

For every input token the pipeline writes a JSON record containing the
two translations, the generated definitions and candidate lemmas, the
top-$K$ synsets with their final and base scores, the lemma-match flag,
and the per-definition contributions. Every prediction is therefore
inspectable, which supports downstream lexicographic reuse: a curator
who disagrees with the top-1 can trace which definition or lemma drove
the result and decide whether to revise the prompt, the gloss, or even the
gold annotation.

\section{Data}
\label{sec:data}

\subsection{A Bilingual Perception-Verb Evaluation Set}

The scarcity of sense-annotated resources for Latin and Ancient Greek is
a recurring obstacle in this line of work. For Latin, one of the only manually
annotated datasets of any size that is mapped to a WordNet inventory is
the Latin portion of the SemEval-2020 Lexical Semantic Change benchmark
\citep{schlechtweg2020semeval, mcgillivray2022new}, whose mapping to LWN
was added \emph{a posteriori}. The PREMOVE dataset
\citep{farina2025premove} contributes around 2{,}800 manually annotated
tokens of Latin and Ancient Greek preverbed motion verbs, with senses encoded as OEWN synsets.
No other comparably curated resources exist.

To support a focused evaluation of \textsc{Inspicio} we built a new
bilingual dataset of 150 perception-verb tokens, with 72 Latin and 78
Ancient Greek instances. The tokens were sampled from the PREMOVE Base
Corpus \citep{farina2025premove}, which is balanced across periods,
genres, and authors for both languages. This balancing carries over to
our sample, which therefore spans Archaic to Late Greek and Early to
Post-Classical Latin.

We targeted perception verbs for three reasons. First, verbs are a particularly polysemous part of speech, which makes them consistently challenging to disambiguate \citep{10.1162/coli_a_00500}.
Second, perception verbs combine concrete sensory meanings with a rich
layer of metaphorical and mental-state extensions: Lat.\ \textit{vide\=o}
and AGr.\ \textit{hor\'a\=o} occur both in literal `see' readings and in
inferential `understand, realise' ones, which makes them a natural stress
test for a pipeline that must separate literal from figurative senses.
Third, perception is a cognitively universal domain whose lexicalisations
are well represented in the OEWN, whereas culturally specific items, often nouns (e.g. the Lat.\ \textit{consul}), have no fitting synset and would force
annotators to invent one or to refuse annotation. Lexical anchors were
drawn from \citet{levin1993english}'s class~30.1, excluding olfactory
and gustatory verbs because of corpus sparsity: Lat.\ \textit{audi\=o} /
AGr.\ \textit{ako\'u\=o}, Lat.\ \textit{vide\=o} / AGr.\ \textit{hor\'a\=o},
Lat.\ prefixed compounds in \textit{-spici\=o} / AGr.\ \textit{bl\'ep\=o},
and Lat.\ \textit{senti\=o} / AGr.\ \textit{aisth\'anomai}.

\paragraph{Annotation.}

Each occurrence was independently annotated by two annotators, both
trained linguists with prior experience in semantic annotation. For
every token, the annotators selected the most appropriate OEWN synset
given the original context, with access to the full sentence and, when
useful, to the surrounding passage. Disagreements were resolved after a round of review to produce a single gold synset per occurrence.
Inter-annotator agreement was computed on the two-annotator
labels using Cohen's $\kappa$ \citep{cohen1960kappa, artstein2008inter}
and reached $\kappa = 0.895$, in the upper range of values typically
reported for fine-grained WordNet annotation \citep{passonneau2010word}.

\subsection{Out-of-Domain Evaluation: PREMOVE}

To test whether the pipeline generalises beyond a curated, semantically
narrow sample, we additionally evaluate on PREMOVE
\citep{farina2025premove}, a manually annotated, diachronic
dataset of Latin and Ancient Greek preverbed motion verbs spanning the
8th century BCE to the 2nd century CE. Its sense annotations are encoded
directly as OEWN synsets, which makes the resource immediately
compatible with our setup. PREMOVE is narrow in scope, covering a single
verb domain, but its tokens are naturalistically distributed: their
frequency profile is Zipfian, metaphorical readings such as
\textit{evenio} `happen' or \textit{invenio} `find' are well represented,
and the genre and authorial range is as wide as that of our curated
sample. It is therefore an informative stress test of a system tuned on
a semantically focused case study.

\subsection{Cross-Lingual Test: Diachronic Italian}

A final evaluation extends \textsc{Inspicio} to a new language and a
different temporal range. We collected 100 Italian motion-verb tokens
from the diachronic MIDIA corpus \citep{iacobini2022corpus}, sampled so
as to match, where possible, the Latin lemmas that appear in PREMOVE
(e.g.\ Lat.\ \textit{incurro} $\rightarrow$ It.\ \textit{incorrere}). The
two sets are therefore comparable at their lexical entry points, but
differ sharply in semantic transparency: Italian descendants of Latin
preverbed motion verbs have undergone substantial lexicalisation and
figurative drift, so a much smaller proportion of their occurrences
retains a literal motion reading. The temporal coverage of the Italian
sample, from late-medieval to contemporary, puts additional pressure on
the LLM's ability to handle diachronic variation in a language for
which pretraining data is unevenly distributed across periods.

Annotation followed the same protocol as for Latin and Ancient Greek,
with two annotators selecting OEWN synsets in context. Cohen's $\kappa$ on the two-annotator labels
reached $0.914$.

\section{Experiments and Results}
\label{sec:experiments}

We evaluate \textsc{Inspicio} on the three datasets introduced in
Section~\ref{sec:data}. The main experiment tests the pipeline on the perception-verb test set across different models, and is designed to address RQ1 and RQ2 jointly.
The PREMOVE and Italian evaluations target RQ4, and a final ablation
study addresses RQ3.

\paragraph{Models and Experimental Setup.}

We pair six instruction-tuned LLMs with six sentence-embedding models.
The LLMs were chosen to cover the current state of the art among
recent open-weights, high-parameter systems that support chain-of-thought reasoning,
an ability that recent work has identified as a strong predictor of
disambiguation accuracy \citep{farina2025premove}:
DeepSeek V3.2\footnote{\url{https://huggingface.co/deepseek-ai/DeepSeek-V3.2}}, DeepSeek V4 Pro\footnote{\url{https://huggingface.co/deepseek-ai/DeepSeek-V4-Pro}}, Kimi K2.6\footnote{\url{https://huggingface.co/moonshotai/Kimi-K2.6}}, GLM 5.1\footnote{\url{https://huggingface.co/zai-org/GLM-5.1}}, Qwen 3.5
397B A17B\footnote{\url{https://huggingface.co/Qwen/Qwen3.5-397B-A17B}}, and Mistral Medium 3.5\footnote{\url{https://huggingface.co/mistralai/Mistral-Medium-3.5-128B}}. All models are queried through
their official API providers with the default decoding parameters
listed in their documentation; the prompts are identical across
systems and are reported in Appendix~\ref{sec:appendix_prompts}. The embedding
models cover both closed and open systems, and were selected to span
different training regimes and parameter counts:
text-embedding-3-large\footnote{\url{https://developers.openai.com/api/docs/models/text-embedding-3-large}},
KaLM-Embedding-Gemma3-12B-2511\footnote{\url{https://huggingface.co/tencent/KaLM-Embedding-Gemma3-12B-2511}}, Qwen3-Embedding-8B\footnote{\url{https://huggingface.co/Qwen/Qwen3-Embedding-8B}}, Cohere Embed v4\footnote{\url{https://cohere.com/blog/embed-4}},
Harrier-OSS-27B\footnote{\url{https://huggingface.co/microsoft/harrier-oss-v1-27b}}, and jina-embeddings-v5-text-small\footnote{\url{https://huggingface.co/jinaai/jina-embeddings-v5-text-small}}. For the embedding models that supported it, we included a short query prompt (see Appendix~\ref{sec:appendix_prompts}). All other
pipeline hyperparameters (Section~\ref{sec:method}) are held fixed
across runs. We report Recall@$k$ at $k \in \{1, 10, 20, 50\}$, with
$k = 50$ as the headline metric.

\paragraph{Perception Verbs.}

\begin{figure*}[t]
    \centering
    \includegraphics[width=0.95\textwidth]{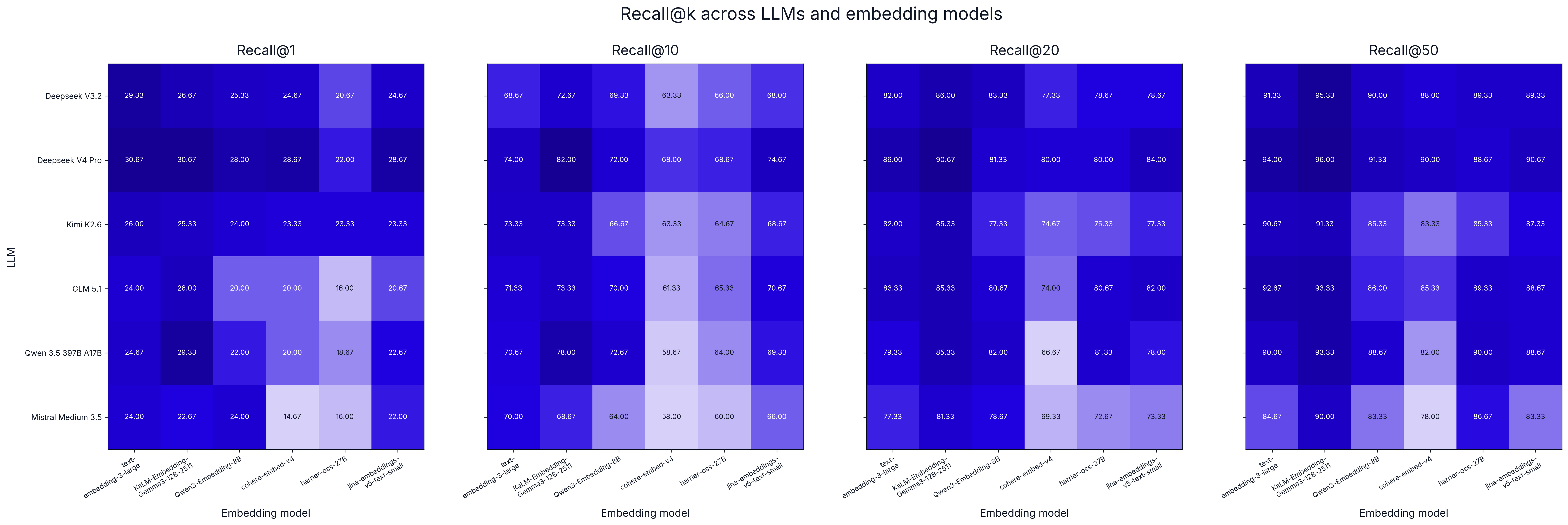}
    \caption{Recall@$k$ heatmaps on the perception-verb test set, for
    $k \in \{1, 10, 20, 50\}$. Rows correspond to LLMs and columns to
    embedding models. Cell shading maps to relative performance within each individual metric, darker colours indicate higher recall.}
    \label{fig:heatmaps}
\end{figure*}

Table~\ref{tab:grid} reports Recall@50 for every $\langle$LLM,
embedding$\rangle$ pair, and Figure~\ref{fig:heatmaps} shows the
corresponding heatmaps at $k \in \{1, 10, 20, 50\}$. The best
combination is DeepSeek V4 Pro with KaLM-Embedding-Gemma3-12B, which
reaches \textbf{96\%} Recall@50. KaLM is the strongest embedding
for five of the six LLMs, while DeepSeek V4 Pro is the strongest LLM
in four of the six embedding columns.


\paragraph{PREMOVE Case Study.}

We then apply the best combination to PREMOVE in order to test whether
performance carries over to a more heterogeneous, naturalistically
distributed sample. Recall@50 reaches 81.65\%
(Table~\ref{tab:supplementary}), roughly 15 points below the
perception-verb result. The gap is consistent with the broader
semantic spread of motion verbs, the Zipfian sense distribution of the
dataset, and the diachronic and metaphorical drift that affects
preverbed forms \citep{farina2025probing}.

\paragraph{Diachronic Italian.}

The same combination is finally evaluated on the diachronic Italian
set (Table~\ref{tab:supplementary}). Recall@50 reaches
\textbf{91\%}, higher than the PREMOVE result. The
outcome is informative given the linguistic distance and diachronic shift involved: most
descendants of Latin preverbed motion verbs have undergone substantial
lexicalisation, and many tokens occur in figurative or
grammaticalised readings that no longer involve literal motion; furthermore, the diachronic stratification of the data means that the tested tokens present widely different meanings across their occurrences in the dataset. Thus, the English-pivot strategy remains effective when the source language is itself well represented in the LLM's pretraining.

\paragraph{Ablation Tests.}

We isolate the contribution of individual pipeline components through
four ablation tests, run with the best combination on the perception-verb
test set (Table~\ref{tab:supplementary}). Removing the translation
stage produces the sharpest drop, to 92\%, indicating
that the literal/natural translation pair contributes information
that the gloss generator alone does not recover. However, such a performance gap may be acceptable in certain cases considering that only a single LLM call (where the model directly identifies candidate definitions and lemmas) is needed, halving the inference cost. Removing the lemma boost causes another drop, to 94\%: dense and sparse retrieval recover synsets that
neither component reaches in isolation, which answers positively to RQ3.  Disabling the definition-rank decay (uniform weights across the three definitions) doesn't seem to decrease performance, as well as disabling MMR. These two parameters are the most subject to model selection and data distribution: MMR can help surface rare senses, which may not always be needed, and definition-rank decay largely depends on the tendency of the LLM to generate relatively similar or markedly different definitions. We retain MMR in the default configuration, since it noticeably diversifies the top of
the ranking without harming aggregate recall, and a diverse top-$K$
is deemed preferable for downstream lexicographic inspection.

\begin{table*}[t]
\centering
\small
\begin{tabular}{lcccccc}
\toprule
& \multicolumn{6}{c}{Embedding model} \\
\cmidrule(lr){2-7}
LLM & OpenAI 3-large & KaLM 12B & Qwen3 8B & Cohere v4 & Harrier 27B & Jina v5 \\
\midrule
DeepSeek V3.2      & 91.33 & 95.33 & 90.00 & 88.00 & 89.33 & 89.33 \\
DeepSeek V4 Pro    & 94.00 & \textbf{96.00} & 91.33 & 90.00 & 88.67 & 90.67 \\
Kimi K2.6          & 90.67 & 91.33 & 85.33 & 83.33 & 85.33 & 87.33 \\
GLM 5.1            & 92.67 & 93.33 & 86.00 & 85.33 & 89.33 & 88.67 \\
Qwen 3.5           & 90.00 & 93.33 & 88.67 & 82.00 & 90.00 & 88.67 \\
Mistral Medium 3.5 & 84.67 & 90.00 & 83.33 & 78.00 & 86.67 & 83.33 \\
\bottomrule
\end{tabular}
\caption{Recall@50 (\%) on the 150-token perception-verb test set,
for every combination of LLM (rows) and sentence-embedding model
(columns). The best cell is highlighted.}
\label{tab:grid}
\end{table*}

\begin{table}[t]
\centering
\small
\begin{tabular}{lcc}
\toprule
Configuration & Recall@50 & Recall@20 \\
\midrule
\multicolumn{2}{l}{\textit{Case Studies}} \\
PREMOVE        & 81.65 & 74.91 \\
Diachronic Italian                     & 91.00 & 84.00 \\
\midrule
\multicolumn{2}{l}{\textit{Ablation Tests}} \\
Full pipeline                          & 96.00 & 90.67 \\
$-$ translation stage                  & 92.00 & 83.33 \\
$-$ definition-rank decay              & 96.00 & 90.00 \\
$-$ lemma boost                        & 94.00 & 85.33 \\
$-$ MMR re-ranking                     & 96.67 & 90.67 \\
\bottomrule
\end{tabular}
\caption{Recall@50 (\%) of the best combination (DeepSeek V4 Pro $+$
KaLM-Embedding-Gemma3-12B-2511) on the out-of-domain (PREMOVE) and
cross-lingual (Italian) test sets, and ablations on the
perception-verb set.}
\label{tab:supplementary}
\end{table}

\section{Discussion}

\subsection{Qualitative Error Analysis}

A close reading of the cases in which the gold synset falls below the
top-$K$ shows a few recurring patterns. The pipeline rarely fails because
the LLM misunderstands the sentence; more often, it produces a
semantically appropriate definition that is nonetheless slightly
different from the gold synset in OEWN's fine-grained sense space. A
representative example is Lat.\ \textit{provide\=o} in a context where
the gold synset is \texttt{oewn-01185006-v}, glossed as ``give what is
desired or needed, especially support, food or sustenance''. The model
returns the definition ``make preparations to meet future needs by
supplying necessary items'', which is paraphrastically close to the
gold gloss but does not surface any of the synset's lemmas
(\textit{cater}, \textit{ply}, \textit{provide}, \textit{supply}). The
candidate lemmas it does propose (\textit{provide for}, \textit{make
provision for}, \textit{take care of}, \textit{see to}, \textit{arrange
for}) all match adjacent synsets, which then occupy higher positions in
the ranking. This kind of failure is largely a granularity artefact of
the type already noted by \citet{bejgu2024word} and
\citet{navigli2026word}, and confirms that the lemma signal is
complementary to the dense one rather than redundant: when both align,
the gold synset surfaces more easily; when only the definition matches, the ranking is dominated by lexically similar neighbours.

Two systematic biases emerged during prompt development and were
addressed before the final evaluation. The first concerns a failure in interpreting co-composition \cite{10.1093/oxfordhb/9780199541072.013.0017}. In preliminary runs, the model tended to incorporate the
semantics of the verb's arguments into the definition of the verb
itself: Lat.\ \textit{iter\=o} in \textit{iter\=o coniugium} (literally
\emph{to repeat a marriage}) was glossed as ``to remarry'', although
\textit{iter\=o} simply means ``to do again'' and the marital reading
arises entirely from \textit{coniugium}. The second concerns negation,
which the model occasionally treated as part of the verb's lexical
content, returning the antonymic sense for negated occurrences. Explicit prompt instructions to keep verb semantics
separate from argument semantics, and to define the verb independently
of any negation in its context, removed both biases on our development
sample. The hybrid scoring scheme makes the pipeline more tolerant to
residual noise of this kind: when the definition is mildly off, lemma
overlap can still pull the correct synset into the top-$K$, and vice
versa.

\subsection{Future Perspectives}

A natural next step is a second-stage reranker: an LLM-based judge
that, given the top-$K$ candidates and the original context, selects
one or two definitive synsets. Comparable architectures have proven
effective for historical languages
\citep{ghizzota2025meaning, farina2025premove, farina2026sense}, and the
high-recall pool produced by \textsc{Inspicio} is well suited to a
reranking stage of this type. The combination would close the gap
between candidate retrieval and final disambiguation, while preserving
the open-vocabulary character of the upstream pipeline.

Once such a reranker is in place, the pipeline can act as a generator
of silver-standard sense annotations for languages that currently lack
any. Annotation propagation through automatic means has already been
shown to complement gold data productively for historical languages
\citep{ghinassi2024language}; an open-vocabulary system would remove the
requirement of an existing sense inventory in the source language. The resulting silver datasets, together with the
intermediate artefacts that \textsc{Inspicio} already logs, could in
turn feed the bootstrapping of dedicated WordNets for these languages,
along the lines of existing efforts behind the Latin and
Ancient Greek WordNets \citep{marchesi2025towards, santoro2025exploring}. The pipeline would thus contribute to building
the very resources whose absence motivates it in the first place. It is important to stress that the pipeline is designed to work in a zero-shot setting, addressing real-world conditions where sense-annotated data may not be available for historical languages. However, supervised fine-tuning strategies may be explored where possible to further improve performance and linguistic understanding. Again, recent advances in agentic LLMs make it an interesting possibility to test the retrieval part of our pipeline in an unconstrained, autonomous setting.

\section*{Limitations}

Our evaluation focuses on verbs, which are a particularly polysemous part of
speech
\citep{10.1162/coli_a_00500}. The architecture is
agnostic to part of speech, and four OEWN indexes are already in place,
but additional evaluations on nouns, adjectives, and adverbs would
strengthen the generality of our findings.

The pipeline relies on LLM predictions at two consecutive stages, and
is therefore stochastic. A misinterpretation produced at translation
time can propagate to gloss generation and, from there, to retrieval.
The translation stage in particular requires a relatively high
sampling temperature, since the literal/natural pair is meant to
expose lexical and stylistic variation. This design choice introduces
a degree of run-to-run variability that more deterministic pipelines
do not face. The auditability of the intermediate outputs partly
mitigates this issue, in that any failure can be traced back to the
specific stage at which it originated, but does not eliminate it.

Finally, routing through English is a deliberate compromise. The OEWN
is the most extensively populated lexical-semantic resource of its
kind \citep{mccrae2020english}, and serves as the pivot of several
bootstrapping projects for historical and low-resource languages
\citep{minozzi2017latin, bizzoni2014making}, but its inventory still
reflects the lexicalisation patterns of contemporary English. Some
source-language senses have no perfect English counterpart, and our
pipeline can at best return the closest available synset in such
cases. As discussed above, the natural mid-term solution is the
incremental construction of dedicated sense inventories for the
languages of interest, a process that \textsc{Inspicio} itself can
help support.

\section*{Acknowledgments}

\bibliography{custom}

\appendix

\section{Prompts}
\label{sec:appendix_prompts}

This appendix presents the complete prompt templates used for all models, formatted in Markdown for clarity and reproducibility. Three prompts are presented: the translation prompt, the definition and lemma generation prompt, and the embedding prompt (only used with models that supported it).

\subsection{Translation Prompt}

\texttt{You are an expert translator specializing in \{language\}.}

\bigskip

\texttt{\# TASK:}
\texttt{Translate the given sentence into English in TWO ways:}

\texttt{1. a literal translation that stays close to the source wording}

\texttt{2. a natural translation that sounds fluent in modern English}

\texttt{Focus on correctly representing the literal and metaphorical meanings of specific words.}

\bigskip

\texttt{\# OUTPUT:}
\texttt{Return ONLY valid JSON with exactly these keys:}

\texttt{\{\{}

\texttt{\ \ "literal": "string",}

\texttt{\ \ "natural": "string"}

\texttt{\}\}}

\bigskip

\texttt{No commentary, no markdown, no extra keys.}

\subsection{Definition and Lemma Generation Prompt}

\texttt{You are an expert lexicographer and linguist specializing in \{language\} semantics.}

\bigskip

\texttt{\# TASK:}

\texttt{You will be given:}

\texttt{- a target token}

\texttt{- its dictionary lemma}

\texttt{- the original sentence it occurs in}

\texttt{- TWO proposed English translations of the sentence - one literal, one natural.}

\texttt{Using all of this context, produce:}

\texttt{1. 1 to 3 possible English dictionary-style definitions of the target word in this context, ordered by likelihood (most likely first). Each definition must be a phrase that fully captures and explains the sense. Example: "she ran all the way home" -> "move rapidly from one place to another"}

\texttt{2. 1 to 5 candidate English lemmas or short expressions (1–2 words, e.g. "make up", "go out") that could translate the token in this context.}

\bigskip

\texttt{\# GUIDELINES:}

\texttt{- Be specific and detailed enough to distinguish senses.}

\texttt{- Account for negation: define the verb's meaning, not its truth value. Example: "she didn't run" -> "move at a speed faster than a walk", not "stand still".}

\texttt{- Account for metaphorical meanings: when in doubt, include both literal and metaphorical definitions. Example: "she saw a risk in his plan" -> "perceive a situation mentally" works better than "perceive by sight".}

\texttt{- Beware of distinguishing the actual meaning of a verb from those of its arguments.}

\texttt{\ \ \ \ - Examples: "He shed a few tears" -> "let fall, emit" is the right choice, not "cry", which would absorb the meaning of "tears".}

\texttt{\ \ \ \ - However, "She threw a party" -> "organize an event" is appropriate, just like "She caught a cold" -> "get struck by an illness", as those are actual meanings conveyed by the verbs.}

\texttt{\ \ \ \ - Avoid unnecessary contextual information: "She ate the cake gleefully" -> "take in food", not "take in food in a joyous manner".}

\texttt{\ \ \ \ - However, "He devoured the cake" -> "eat quickly and hungrily" is correct, because the specific manner of eating is a core semantic feature inherently lexicalized in the verb itself.}

\texttt{- If there is genuine ambiguity, include multiple definitions; otherwise output 1.}

\texttt{- Keep outputs precise and consistent.}

\bigskip

\texttt{\# OUTPUT:}

\texttt{Return ONLY valid JSON with exactly these keys:}

\texttt{\{\{}

\texttt{\ \ "definitions": ["def1", "def2", "def3"],}

\texttt{\ \ "candidate\_lemmas": ["lemma1", "lemma2", "lemma3", "lemma4", "lemma5"]}

\texttt{\}\}}

\bigskip

\texttt{No commentary, no markdown, no extra keys.}

\subsection{Embedding Prompt}

\texttt{Given a dictionary definition, retrieve the WordNet synset that best matches its meaning.}

\end{document}